\documentclass{article} 
\PassOptionsToPackage{numbers, compress}{natbib}
\pdfoutput=1
\usepackage[final]{neurips_2019_ml4ps}
\setcitestyle{square, comma, numbers,sort&compress, super}
\usepackage{url}
\usepackage{xcolor}
\usepackage{multirow}
\usepackage{booktabs}

\title{Unsupervised Star Galaxy Classification with Cascade Variational Auto-Encoder}

\author{%
Hao Sun\textsuperscript{1,2}\thanks{Author is now affiliated with CUHK. This work was done in the visiting research intern program in the LCDM group at UIUC},
Jiadong Guo\textsuperscript{2},
Edward J. Kim\textsuperscript{3},
Robert J. Brunner\textsuperscript{3}\\
\textsuperscript{1}Peking University,
\textsuperscript{2}Peng Cheng Laboratory,
\textsuperscript{3}The University of Illinois at Urbana Champaign
}

\usepackage{amsmath}
\usepackage{amssymb}
\usepackage{hyperref}
\usepackage{graphicx}
\usepackage{subfigure}

\begin{document}
\maketitle
\begin{abstract}
The increasing amount of data in astronomy provides great challenges for machine learning research. Previously, supervised learning methods achieved satisfactory recognition accuracy for the star-galaxy classification task, based on manually labeled data set. In this work, we propose a novel unsupervised approach for the star-galaxy recognition task, namely Cascade Variational Auto-Encoder (CasVAE). Our empirical results show our method outperforms the baseline model in both accuracy and stability.
\end{abstract}




\section{Introduction}
\label{intro}
A main challenge in astronomical photometric surveys, e.g. the Sloan Digital Sky Survey (SDSS) \cite{adelman2006fourth}, the Dark Energy Survey (DES) \cite{dark2005dark} and the Large Synoptic Survey Telescope (LSST) \cite{kaiser2002pan}, is the need for object recognition. Their growing scale of collected data in current and future research makes it impossible for human experts to do classification manually. In previous work, machine learning and deep learning techniques were introduced to tackle the challenge \cite{article4,Kim2017star,sevilla2015effect,fadely2012star,kim2015hybrid,kamdar2015machine}, while all of those methods rely on data sets with labels assigned by experts. 

The main drawbacks of previous learning approaches are twofold. On the one hand, labeling astronomical data is a time-consuming, interminable and error-prone job. On the other hand, the label process introduces prior knowledge into the data set, and cognition bias inevitably results. Learning through unsupervised algorithms can be a substitute but we need to extract useful features for the task; and, in most cases, specific network architecture design or domain knowledge is needed to achieve better task specific performance \cite{ronneberger2015u,long2015fully,dosovitskiy2015flownet,he2017mask}. Applying previous unsupervised learning methods into in the domain of astronomy is not straightforward. Specifically, focusing on the star-galaxy classification task, prevailing unsupervised methods: the Auto-Encoders (AEs) and Variational Auto-Encoders (VAEs) \cite{bourlard1988auto,kingma2013auto}, are not able to provide satisfactory results for they are designed as generative models, which will learn the most useful hidden representation for identity mapping from the input space to the output space. Other traditional approaches like t-SNE and ISOMAP are not suitable to cope with large scale image inputs \cite{maaten2008visualizing,balasubramanian2002isomap}, thus combination of AEs and manifold learning methods are used \cite{pratiher2018manifold,vincent2008extracting}.

In this work, we propose a method to produce accurate star-galaxy classification by learning directly from the pixel values of photometric images. Our proposed method, Cascade Variational Auto-Encoder, improves the vanilla VAE \cite{kingma2013auto} to be more capable in classification tasks. We demonstrate our method on the SDSS data set that results in remarkable accuracy improvements over baseline method.

\section{Method}
In this section we will introduce our proposed solution for the unsupervised star-galaxy classification task. The dataset we used will be introduced in Sec.\ref{sec_dataset}. Sec.\ref{sec_math} provides a detailed mathematical formalization of the problem, and proposes an improvement over the prior of vanilla VAEs. We discuss some baseline methods in Sec.\ref{pre}, and finally present our solution in Sec.\ref{sec_casvae}.
\subsection{Dataset}
\label{sec_dataset}

\begin{figure}[t]
\subfigure[Labeled data set]{
\begin{minipage}[t]{1.0\linewidth}
\centering
\includegraphics[width=1.5in]{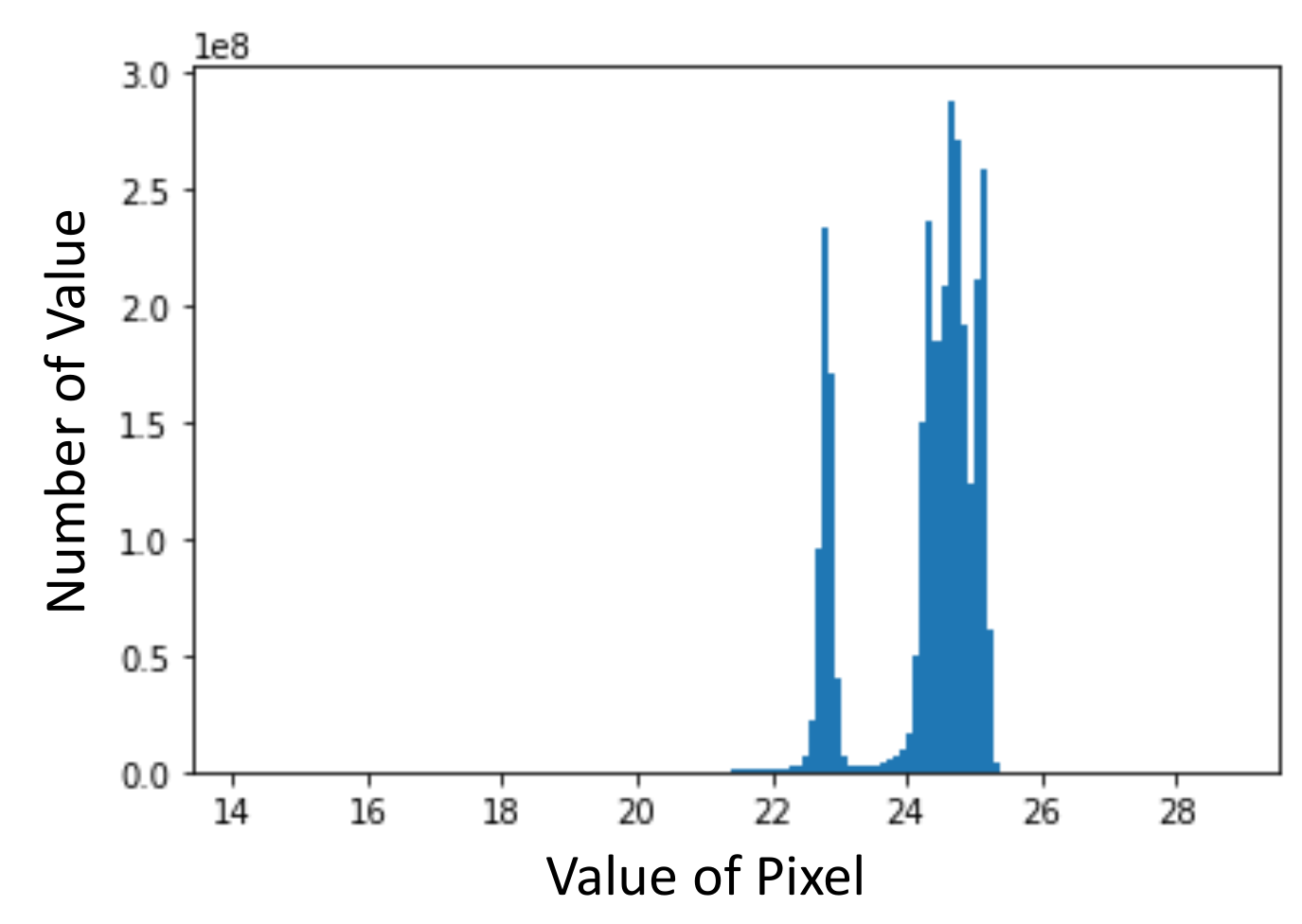}
\includegraphics[width=3.8in]{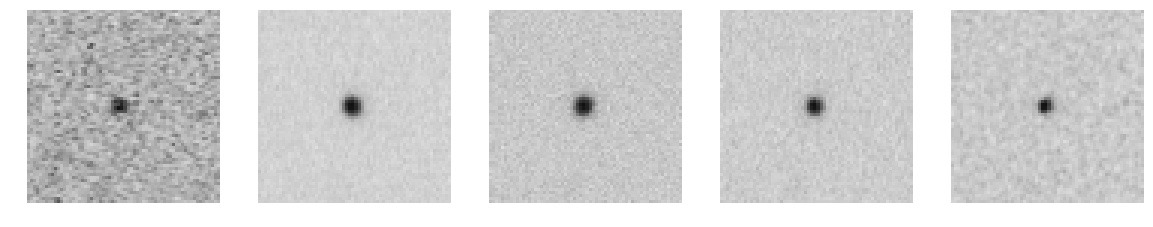}
\label{fig2:side:a}
\end{minipage}%
}
\subfigure[Unlabeled data set]{
\begin{minipage}[t]{1.0\linewidth}
\centering
\includegraphics[width=1.5in]{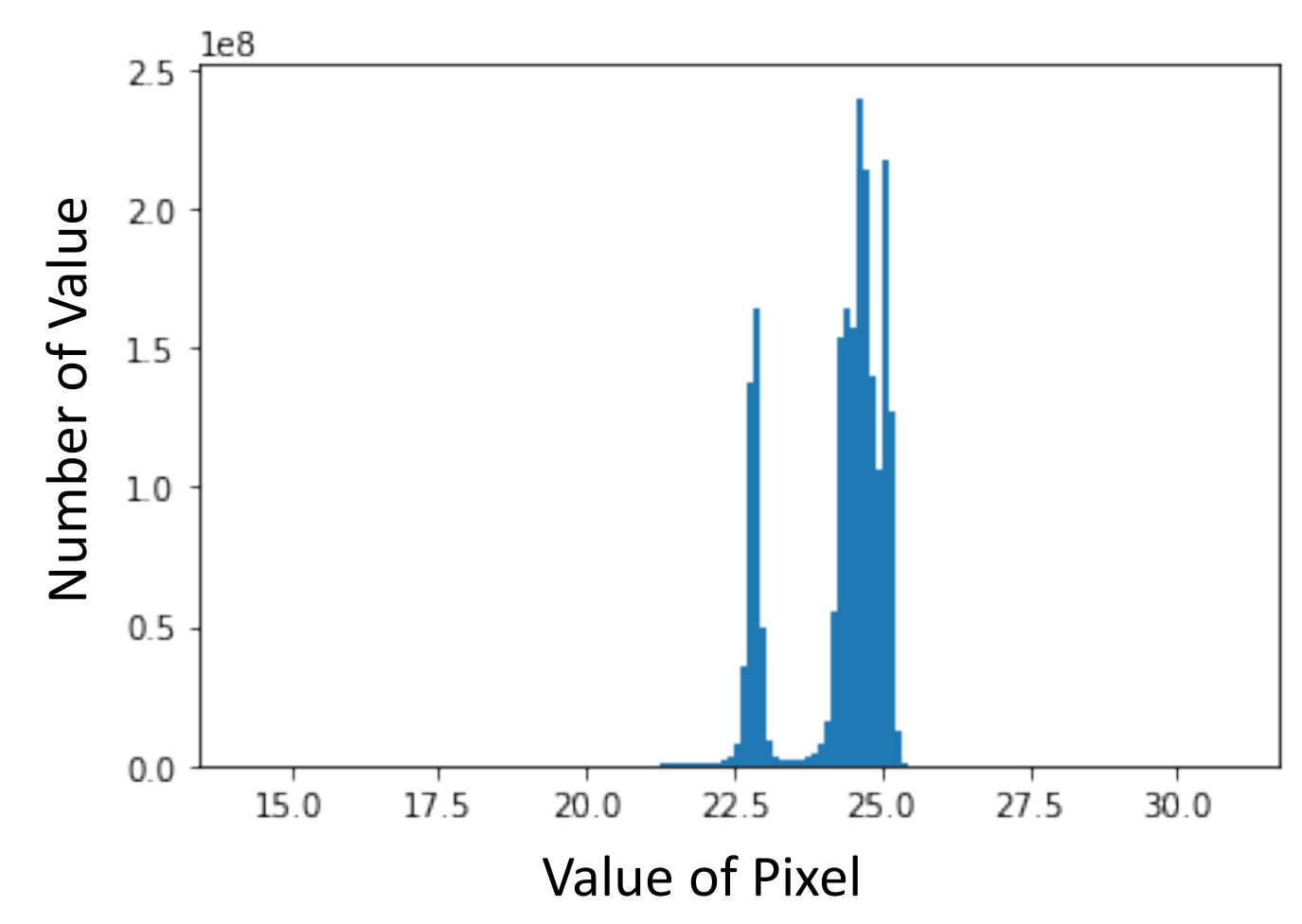}
\includegraphics[width=3.8in]{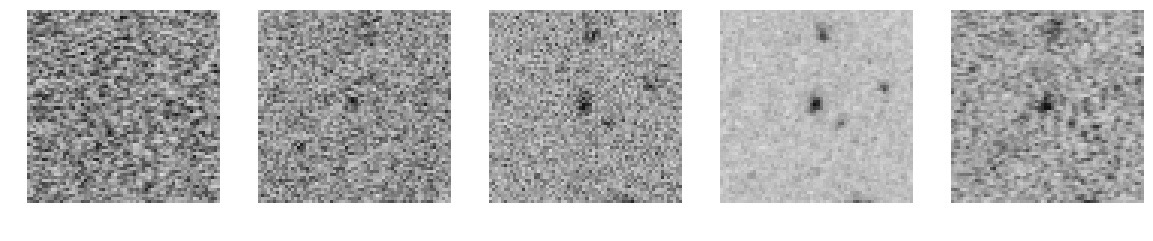}
\label{fig2:side:b}
\end{minipage}
}
\caption{histograms of pixel values and some image samples}
\label{hist_img}
\end{figure}

The dataset we choose to use in this work is the SDSS dataset. Specifically, we have a labeled subset $X_{l}$ and an unlabeled subset $X_{u}$. The labeled subset $X_{l}$ contains 140,000 images, each of which has the shape of $64\times 64\times 5$, the $5$ channels correspond to the $u,g,r,i,z$ bands separately. The ground truth of their class, i.e. stars or galaxies, are also recorded in $X_l$. Further information of the objects like red-shift $z$ and average magnitude are also recorded but not used in our training or evaluation process. In the unlabeled subset $X_{u}$, there are 100,000 images that share the same size with $X_{l}$. But we don't have ground truth labels for those images. We aim to train our model with $X_{u}$ and evaluate our method on $X_{l}$. Fig. \ref{hist_img} shows the histograms of pixel values and some image samples of $X_{l}$ and $X_{u}$ before normalization.

\subsection{Mathematical Formalization}
\label{sec_math}
The mathematical formulation of our task is related to the work of Model Agnostic Meta-Learning (MAML) \cite{finn2017model} but we concentrate more on unsupervised classification tasks instead of meta-learning tasks. Unsupervised classification tasks can be interpreted as functional optimization problems when we aim to train the model with a labeled dataset and validate on the un-labeled dataset. In such problems, we have an labeled data set $X_{l}$, their corresponding labels $Y_{l}$ and the unlabeled data set $X_{u}$. The optimization objective is
\begin{equation}
\label{eq_primal}
\min_f \parallel Y_{l} - f(X_{l})\parallel^2
\end{equation}
where $f$ is a classifier function that maps the original $64\times 64\times 5$ input images into a one dimensional scalar, i.e. the predicted label. As we can only learn from unlabeled data in unsupervised cases, we can only utilize the following objective in our optimization:

\begin{equation}
\label{eq_surro}
\min_{\mathcal{L},f} \mathcal{L}(X_{u},f(X_{u}))
\end{equation}

where $f$ is the classifier that we will use to test our classification model on $X_l$, and $\mathcal{L}$ is a surrogate loss function to be determined, which varies between different methods. Ideally, optimization towards the direction of minimizing the surrogate loss function $\mathcal{L}$ should drive $f$ to be a good classifier automatically. 

With such formalization, the crux is to find an appropriate surrogate loss function $\mathcal{L}$, with which the optimization goal of Eq.(\ref{eq_surro}) can approximate the goal of Eq.(\ref{eq_primal}). Previous work regard AEs and VAEs as dimension reduction models and utilize reproduction error as the surrogate loss $\mathcal{L}$, during which the bottleneck structure of AEs or VAEs can benefit classification due to its low dimension \cite{betechuoh2006autoencoder,geng2015high,sun2016sparse}. However, vanilla VAEs take a Gaussian prior in its hidden space to facilitate computation, and enables generating a consistent image by adding perturbations into the hidden variables before putting it to the decoder networks. Noticing the $\mathcal{N}(0,1)$ prior is not conducive to unsupervised classification, Dilokthanaku et al. put forward GMVAE \cite{dilokthanakul2016deep}, but their work can not scale up to large scale input images. In our work, as we are interested in separating stars from galaxies, we need to do two-class object recognition, and a better and more convenient choice is to leverage a double-peak Gaussian prior to replace the vanilla Gaussian prior in VAEs. Denoting the prior of hidden space as $P(z)$, we use

\begin{equation}
\label{eq_dual}
P(z)\sim \alpha \mathcal{N}(-\mu,\sigma^2)+(1-\alpha)\mathcal{N}(\mu,\sigma^2)
\end{equation}

where $\alpha$ is a weighting factor. Although we can choose to use the proportion of stars and galaxies as a prior, we choose to use $\alpha = \frac{1}{2}$ in our experiments to avoid a dependence on prior knowledge.

\subsection{Baseline Methods}
\label{pre}
In unsupervised clustering, AEs are always utilized to reduce the dimension of input data and Manifold Learning (ML) methods (e.g. ISOMAP, t-SNE) are applied to perform further dimension reduction and classification \cite{jiang2016variational,badino2014auto,vincent2008extracting,pratiher2018manifold}. In detail, those methods have three steps. First, such methods utilize VAEs or AEs to extract features from original images and reduce the dimensions into $dim_{hid}$, where $dim_{hid}\ll dim_{input}$. Next, they run an ML algorithm to map the $d$ dimensional $dim_{hid}$ into a one dimensional scalar. Finally, the last step is to use the scalar and a threshold, which can be determined from prior knowledge like star-galaxy proportion, to perform classification. In our experiments we will use the method of combining VAEs and ML as baseline.

\begin{figure}[tbp!]
\centering
\includegraphics[width=5.5in]{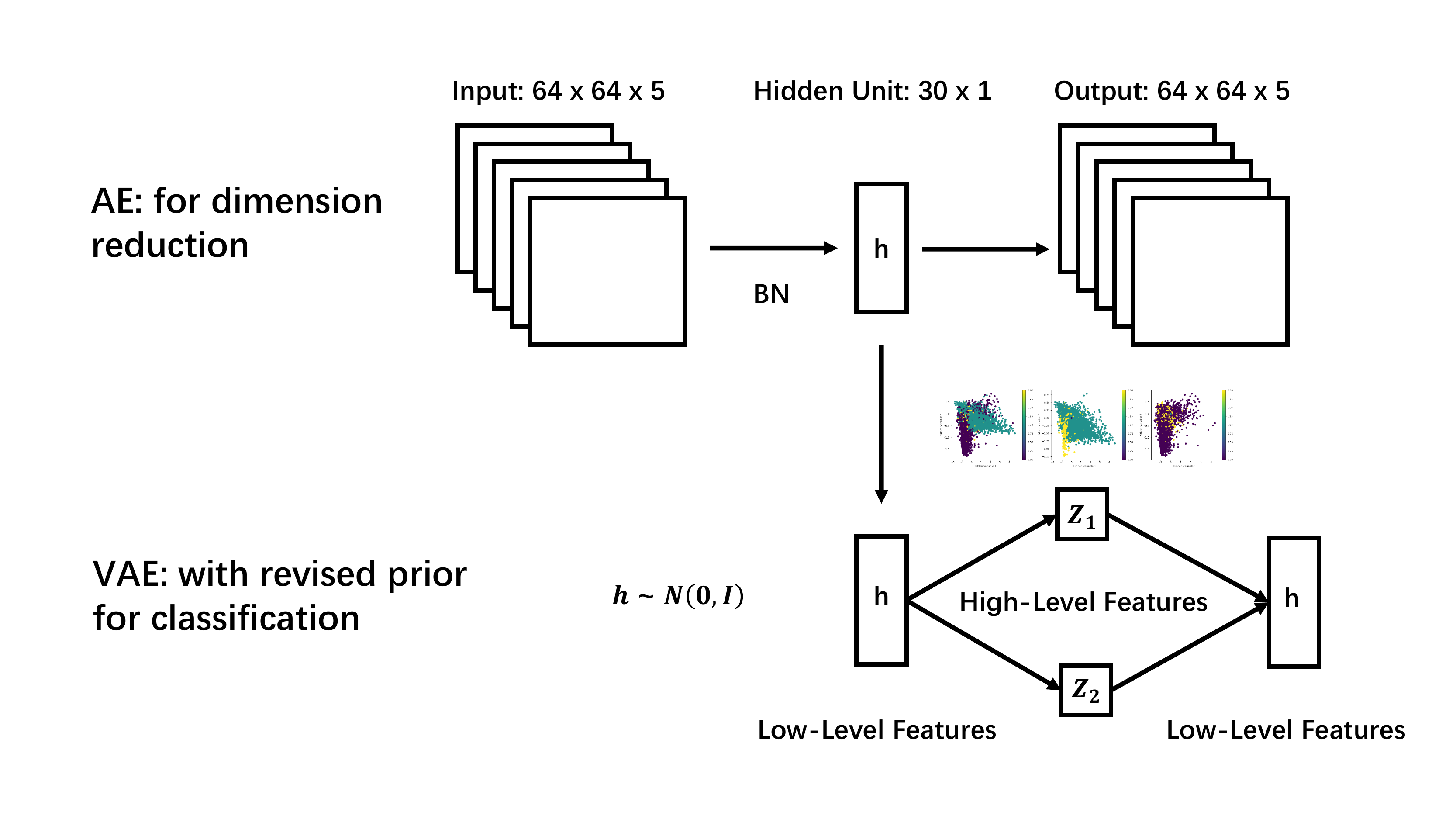}
\caption{the learning paradigm of our proposed method. Our method can be interpreted as a VAE with its reproduction loss defined by an AE}
\label{pipeline}
\end{figure}

\subsection{Cascade VAE}
\label{sec_casvae}
Our proposed method, namely Cascade VAE (CasVAE), introduces a hierarchical structure based on VAEs to facilitate unsupervised classification. Based on previous work, most successful generative models often use only a single layer of latent variables \cite{radford2015unsupervised,van2016conditional}, and multiple layers only show modest performance increases in quantitative metrics such as log-likelihood \cite{sonderby2016ladder,bachman2016architecture}. We choose to use shallow neural networks in our model and separating the dimension reduction process and the classification process.

The modelling ability of VAEs varies when using different number of hidden units. While more hidden units lead to sufficient representation ability, less hidden units prompts higher level features, which is more useful for classification \cite{doersch2016tutorial}. To address this trade-off problem, we proposed a cascade structure. In the first phase of our model, we focus on the dimension reduction part of unsupervised classification, where a normal AE with $30$ hidden units is utilized. We apply batch normalization (BN) \cite{ioffe2015batch} in AE to bridge the gap between different data distributions. The effect of AE in our model can be interpreted as an low-level feature extractor. In the second phase, we utilize a VAE with $2$ hidden units in its hidden layer for classification. The first of the two hidden units follows the vanilla VAE to use a simple Gaussian prior and is used to store information useful for VAE-reproduction. The second hidden unit is equipped with a revised Double-peak Gaussian prior to encourage separation. Fig. \ref{pipeline} shows our architecture, which can be interpreted as a two-channel VAE with its loss function defined by an AE.

\section{Experiments}
\label{exp}
\begin{figure}[tbp!]
\label{fig_result}
\subfigure[VAE+ML]{
\begin{minipage}[t]{0.3\linewidth}
\centering
\includegraphics[width=1.28in]{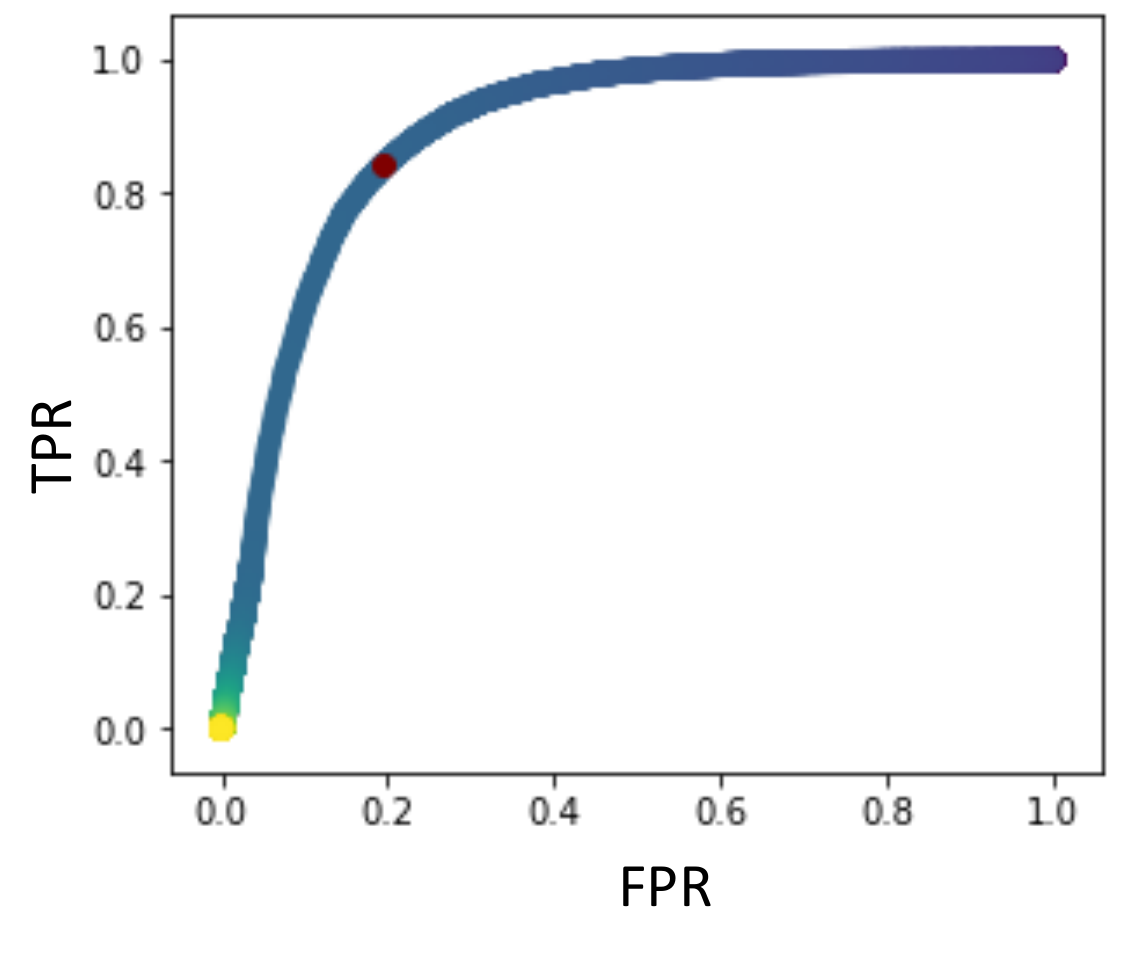}
\label{fig1:side:a}
\end{minipage}%
}
\subfigure[DKL-VAE + ML]{
\begin{minipage}[t]{0.3\linewidth}
\centering
\includegraphics[width=1.28in]{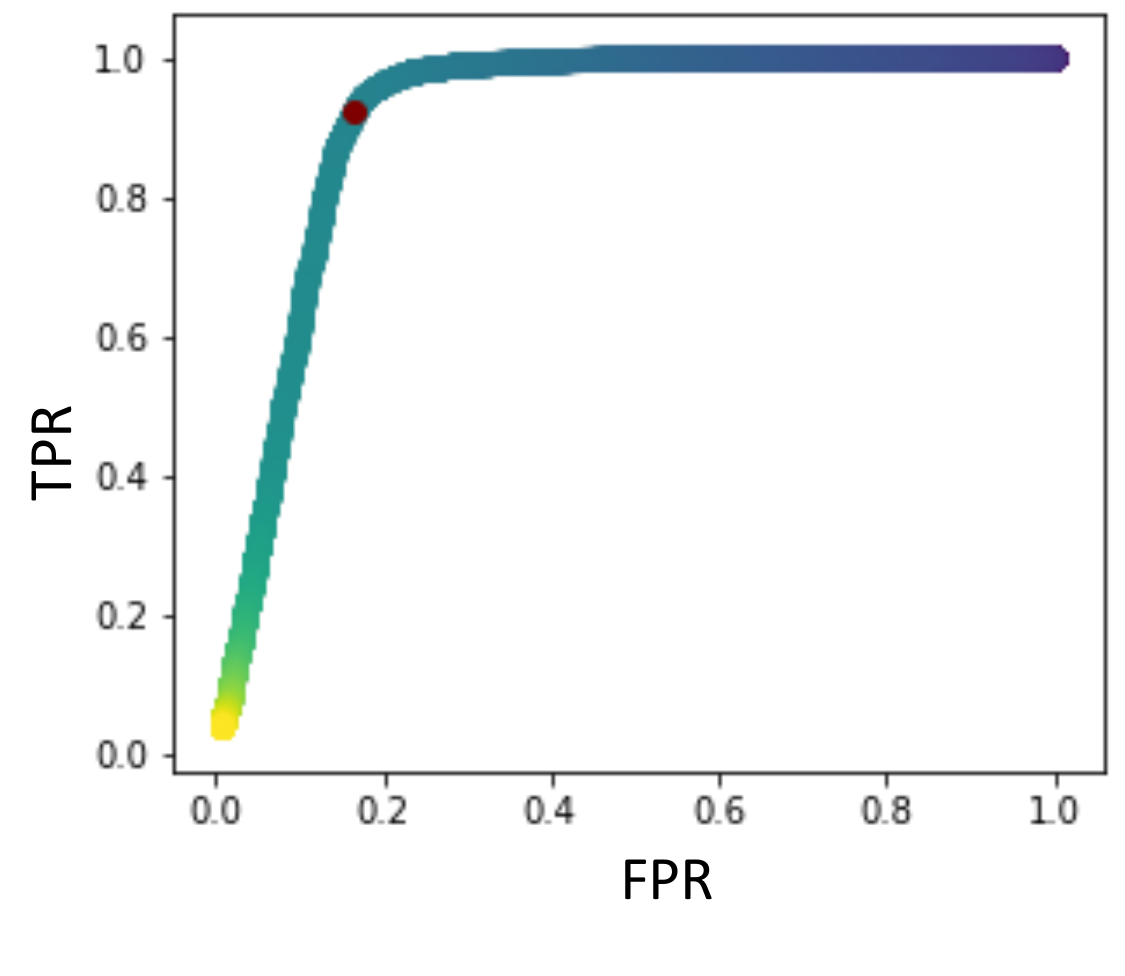}
\label{fig1:side:c}
\end{minipage}
}
\subfigure[CasVAE]{
\begin{minipage}[t]{0.3\linewidth}
\centering
\includegraphics[width=1.28in]{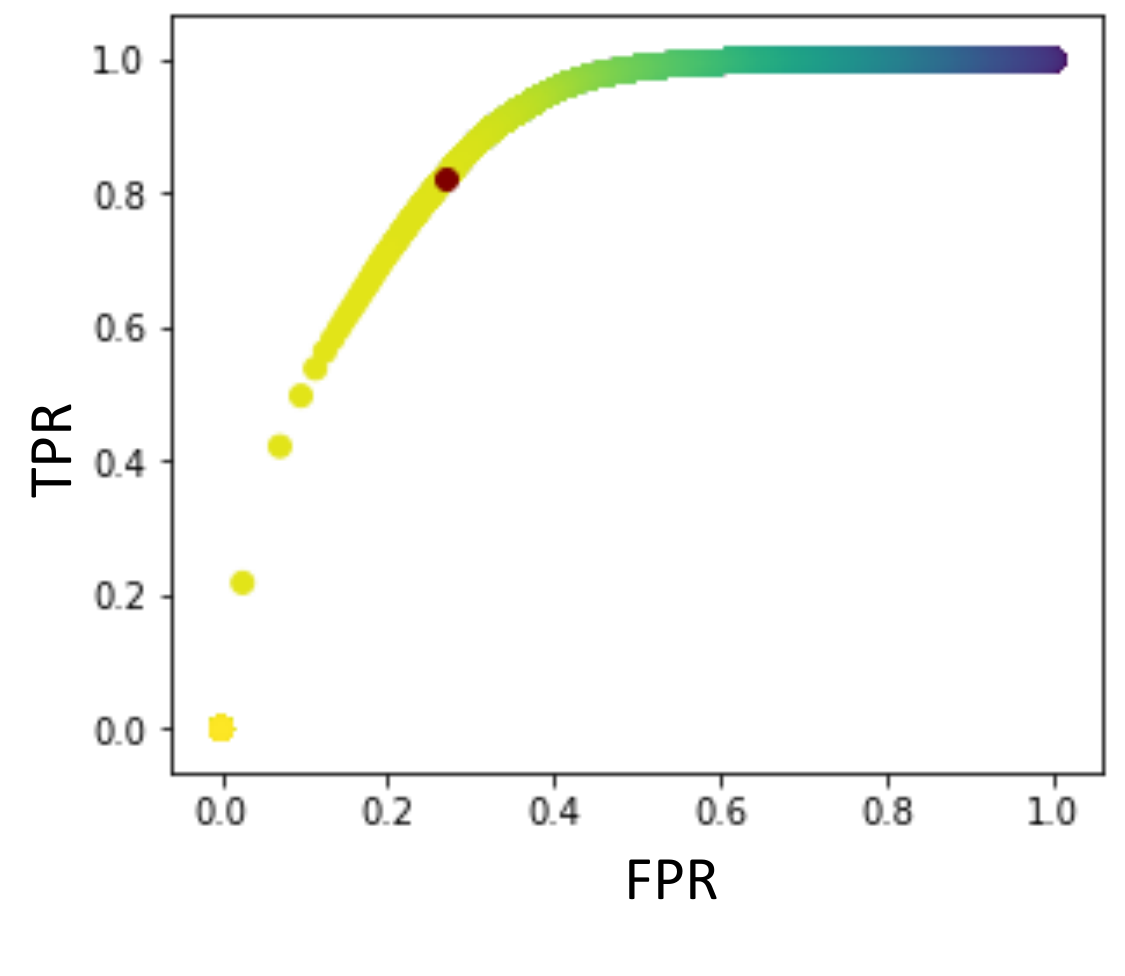}
\label{fig1:side:d}
\end{minipage}
}
\caption{ROC and AUC of each method, (a) VAE + ML as baseline; (b) VAE with double-peak Gaussian prior + ML; (c) Cascade VAE. Points with different colors shows different thresholds (dark/blue color shows smaller threshold and light color shows larger threshold). The red points in each figure shows the point closest to the $(0,1)$ point}
\end{figure}
\begin{table}[t]
\caption{Performance comparison of each method, VAE + ML is the baseline method and DKL-VAE +ML is based on our improvement of the Gaussian prior used in VAE. Results are averaged by $100$ experiments}
\label{tab_1}
\begin{center}
\begin{tabular}{llll}
\toprule \multicolumn{1}{c}{\bf Method}  &\multicolumn{1}{c}{\bf Mean AUC} &\multicolumn{1}{c}{\bf Highest AUC} &\multicolumn{1}{c}{\bf Lowest AUC}\\ \midrule
VAE + ML         &0.75  &0.89 & 0.61\\
DKL-VAE + ML          &0.84 &0.92 &0.74 \\
CasVAE          &\textbf{0.90} &\textbf{0.91} &\textbf{0.90} \\
\bottomrule
\end{tabular}
\end{center}

\end{table}
We approximate the calculation of KL-divergence in VAEs in several ways after we introduce the double-peak Gaussian prior in Eq.(\ref{eq_dual}). We evaluate our method based on 4 different kinds of substitutes of KL-divergence. Those loss terms are double-peak KL-Scaling (DKLSC), double-peak KL (DKL), Wasserstein loss (W) and pseudo Wasserstein loss (PW). More details on the calculation can be found in Supplementary materials

In our experiment, we use the Receiver Operating Characteristic curve (ROC) to visualize the classification results and use Area Under Curve (AUC) to evaluate the performance of our models.
We use grid search for the selection of hyper-parameters and repeat each experiment with different initializations to test the stability of each method. Fig. \ref{fig_result} shows some of our results, We first tried VAE + Manifold Learning (VAE+ML) and show the ROC curve in Figure Fig. \ref{fig_result}(a). Fig. \ref{fig_result}(b) shows our results with a double-peak Gaussian prior. Finally, Fig. \ref{fig_result} (c)  shows our result when we use the Cascade VAE method (CasVAE). 

The first two methods are quite sensitive to the initialization of the neural networks in our experiment. The instability mainly comes from the manifold learning algorithm we use after VAEs. Specifically, Manifold learning algorithms are often sensitive to the outliers. While there are many multi-object images in our dataset, i.e., stars and galaxies may appear in the same image, the prediction we hope to make is on the class of the object in the center of the image. These mixture images may be classified according to the central objects in supervised learning by attention mechanism \cite{vaswani2017attention}. But in unsupervised learning and especially in manifold clustering, they are outliers. Our CasVAE method uses one hidden unit as reproducer and one hidden unit as classifier, so that we can use the classifier unit to do classification instead of using ML in other approaches. The performance of each method in terms of AUC is shown in Table \ref{tab_1}, indicating the excellence performance of our method as well as its stability.

\section{Conclusion}
In this work, we proposed a new approach for unsupervised star-galaxy classification, namely Cascade Variational Auto-Encoder (CasVAE). We highlight two improvements over vanilla VAEs. We first introduce an AE in CasVAE to perform dimension reduction, simplifying the problem to a great extent. Moreover, our revision of the Gaussian prior in vanilla VAEs enables CasVAE to abstract high level features that can be used to perform classification in its inner phase. Compared with previous approaches, CasVAE is able to achieve the highest star-galaxy classification accuracy with high stability, which ensures the reproducibility of our proposed method in applications. 

\textbf{Acknowledgement:} This work was supported and funded by School of Physics of Peking University (The Undergraduate Student Summer Research Intern Fund). We would like to acknowledge discussions with Binxu Wang, Feng Cheng and Dogezi Song.

\newpage
\appendix
\section*{Appendix}
\section{Detailed calculation of KL-divergence}
\subsection{An scaling technique}
\begin{align*}
&D_{KL}\left(N(\mu,\sigma^2) \Vert \frac{1}{2}N(-m,s^2)+\frac{1}{2}N(m,s^2)\right)\\
=&D_{KL}\left(2*\frac{1}{2}N(\mu,\sigma^2)\Vert \frac{1}{2}N(-m,s^2)+\frac{1}{2}N(m,s^2)\right)
\\
\leq & D_{KL}\left(\frac{1}{2}N(\mu,\sigma^2)\Vert \frac{1}{2}N(-m,s^2)\right)+D_{KL}\left(\frac{1}{2}N(\mu,\sigma^2)\Vert \frac{1}{2}N(m,s^2)\right)
\\
=& -\frac{1}{2}\log{2}\left( s^2 + \log{\sigma^2} -\frac{1}{2}(\mu-m)^2 -\frac{1}{2}(\mu+m)^2 -\sigma^2 \right)
\end{align*}
\subsection{Exact solution}

\begin{align*}
&D_{KL}\left(N(\mu,\sigma^2) \Vert \frac{1}{2}N(-m,s^2)+\frac{1}{2}N(m,s^2)\right)\\
=&\int_{-\infty}^{\infty}N(\mu,\sigma^2) \log{\frac{N(\mu,\sigma^2) }{\frac{1}{2}N(-m,s^2)+\frac{1}{2}N(m,s^2)}}dx\\
=&\int_{-\infty}^{\infty}\frac{1}{\sqrt{2\pi}\sigma}e^{-\frac{(x-\mu)^2}{2\sigma^2}}\log{\frac{\frac{1}{\sigma}e^{-\frac{(x-\mu)^2}{2\sigma^2}}}{\frac{1}{2s}[e^{-\frac{(x-m)^2}{2s^2}}+e^{-\frac{(x+m)^2}{2s^2}}]}}dx\\
=& \int_{-\infty}^{\infty}\frac{1}{\sqrt{2\pi}\sigma}e^{-\frac{(x-\mu)^2}{2\sigma^2}}\log{\frac{2s}{\sigma}}dx \\
-& \int_{-\infty}^{\infty}\frac{1}{\sqrt{2\pi}\sigma}e^{-\frac{(x-\mu)^2}{2\sigma^2}}[\frac{(x-\mu)^2}{2\sigma^2}-\frac{(x-m)^2}{2s^2}] dx \\
-& \int_{-\infty}^{\infty}\frac{1}{\sqrt{2\pi}\sigma}e^{-\frac{(x-\mu)^2}{2\sigma^2}}\log{[1+e^{-\frac{2mx}{s^2}}]}dx\\
=& \alpha - \beta - \gamma \\
&\quad \\
\alpha &= \log{\frac{2s}{\sigma}}\\
\beta &= -\frac{(m-\mu)^2+\sigma^2 - s^2}{2s^2}\\
\gamma &\approx -\frac{2m[-\sigma e^{-\frac{\mu^2}{2\sigma^2}}+\sqrt{\frac{\pi}{2}}\mu Erfc(\frac{\mu}{\sqrt{2}\sigma})]}{s^2}
\end{align*}
with an approximation of $Erfc(x) \approx 1-tanh(1.19x)$
\begin{align*}
& D_{KL}\left(N(\mu,\sigma^2) \Vert \frac{1}{2}N(-m,s^2)+\frac{1}{2}N(m,s^2)\right) \\
\approx &\log{\frac{2s}{\sigma}}+\frac{(m-\mu)^2+\sigma^2 - s^2}{2s^2} \\
+&\frac{2m\{-\sigma e^{-\frac{\mu^2}{2\sigma^2}}+\sqrt{\frac{\pi}{2}}\mu [1-tanh(1.19\frac{\mu}{\sqrt{2}\sigma})]\}}{s^2}
\end{align*}

\end{document}